\documentclass[conference]{IEEEtran}
\usepackage{times}
\usepackage{epsfig}
\usepackage{graphicx}
\usepackage{amsmath}
\usepackage{amssymb}
\usepackage{lwm}
\usepackage{url}
\usepackage[numbers]{natbib}
\usepackage{algorithm}
\usepackage{algorithmic}
\usepackage{epsfig}
\usepackage{verbatim}
\usepackage{subcaption}
\usepackage{macros}
\usepackage{xcolor}
\usepackage{color}
\usepackage{listings}

\lstdefinelanguage{Scala}{
  morekeywords={abstract,case,catch,class,def,%
    do,else,extends,false,final,finally,%
    for,if,implicit,import,match,mixin,%
    new,null,object,override,package,%
    private,protected,requires,return,sealed,%
    super,this,throw,trait,true,try,%
    type,val,var,while,with,yield},
  otherkeywords={=>,<-,<\%,<:,>:,\#,@},
  sensitive=true,
  morecomment=[l]{//},
  morecomment=[n]{/*}{*/},
  morestring=[b]",
  morestring=[b]',
  morestring=[b]"""
}

\definecolor{dkgreen}{rgb}{0,0.6,0}
\definecolor{gray}{rgb}{0.5,0.5,0.5}
\definecolor{mauve}{rgb}{0.58,0,0.82}

\usepackage{float}

\begin{document}
\title{\mlint: An API for Distributed Machine Learning}
\author{
Evan R. Sparks\textsuperscript{a} \quad Ameet Talwalkar\textsuperscript{a} \quad Virginia Smith\textsuperscript{a}
\quad Jey Kottalam\textsuperscript{a}
\\
Xinghao Pan\textsuperscript{a} \quad Joseph Gonzalez\textsuperscript{a} \quad Michael J. Franklin\textsuperscript{a} \quad Michael I. Jordan\textsuperscript{a} \quad Tim Kraska\textsuperscript{b} \\
\noindent\textsuperscript{a}University of California, Berkeley \quad 
\noindent\textsuperscript{b}Brown University \\ 
\tt\footnotesize{\{sparks, ameet, vsmith, jey, xinghao, jegonzal, franklin, jordan\}@cs.berkeley.edu, kraskat@brown.edu}
}

%

\newcommand{\tim}[1]{\textcolor{blue}{TIM: #1}}
\newcommand{\joey}[1]{\textcolor{cyan}{JOEy: #1}}
\newcommand{\ameet}[1]{\textcolor{red}{[Ameet: #1]}}
\newcommand{\evan}[1]{\textcolor{green}{EVAN: #1}}
\maketitle

\begin{abstract}
\mlint is an Application Programming Interface designed to address the
challenges of building Machine Learning algorithms in a distributed setting
based on data-centric computing. Its primary goal is to simplify the
development of high-performance, scalable, distributed algorithms. Our initial
results show that, relative to existing systems, this interface can be used to
build distributed implementations of a wide variety of common Machine Learning
algorithms with minimal complexity and highly competitive performance and
scalability.
\end{abstract}

%
\IEEEpeerreviewmaketitle

\section{Introduction} 

The recent success stories of machine learning (ML) driven applications
have created an increasing demand for scalable
ML solutions.  Nonetheless, ML researchers often prefer to code their solutions
in statistical computing languages such as MATLAB or R, as these languages
allow them to code in fewer lines using syntax that resembles high-level
pseudocode. MATLAB and R allow researchers to avoid low-level implementation
details, leading to quickly developed prototypes that are often sufficient for small scale
exploration. 
However, these prototypes are typically 
ad-hoc, non-robust, and non-scalable implementations.
In contrast, industrial implementations of these solutions often require a
relatively heavy amount of development effort and are difficult to change once
implemented. 

This disconnect between these ad-hoc scripts and the growing need for scalable
ML, in particular systems that leverage the increasingly pervasive cloud
computing architecture, has spurred the development of several distributed
systems for ML.  Initial attempts at such systems exposed a restricted set of
low-level primitives for development, e.g., MapReduce~\cite{mahout} or
graph-based~\cite{Low10,Gonzalez12} interfaces. The resulting systems are indeed
significantly faster and more scalable than MATLAB or R scripts.  They also
tend to be much less accessible to ML researchers, as ML algorithms do not
always naturally fit into the exposed low-level primitives, and moreover,
efficient use of these primitives requires a fairly advanced knowledge of the
underlying distributed system. 

Subsequent attempts at distributed systems have
exposed high-level interfaces that compile down to low-level
primitives.  These systems abstract away much of the communication and
parallelization complexity inherent in distributed ML implementations. Although
these systems can in theory obtain excellent performance, they are quite
difficult to implement in practice, as they either heavily rely on optimizers
to effectively transform high-level code into efficient distributed
implementations~\cite{SystemML,HyracksDatalog}, or utilize pattern matching techniques
to identify regions that can be replaced by low-level
implementations~\cite{optiml}. 
The need for fast ML algorithms
has also led to the development of highly specialized systems for ML
using a restricted set of algorithms~\cite{Shogun,vw}, with varying degrees of 
scalability.


Given the accessibility issues of low-level systems and the implementation
issues of the high-level systems, ML researchers have yet to widely adopt any
of the existing systems.  Indeed, ML researchers, both in academic and industrial
environments, often rely on system programmers to translate the prototypes of
their novel, and often subtle, algorithmic insights into scalable and robust
implementations. Unfortunately, there is often a `loss in translation' during
this process; small misinterpretation and/or minor errors are unavoidable and
can significantly impact the quality of the algorithm.  Furthermore, due to the
randomized nature of many ML algorithms, it is not always straightforward to
construct appropriate test-cases and discover these bugs.

\begin{figure}[ht!]
\centering
  \ipsfig{.45}{figure=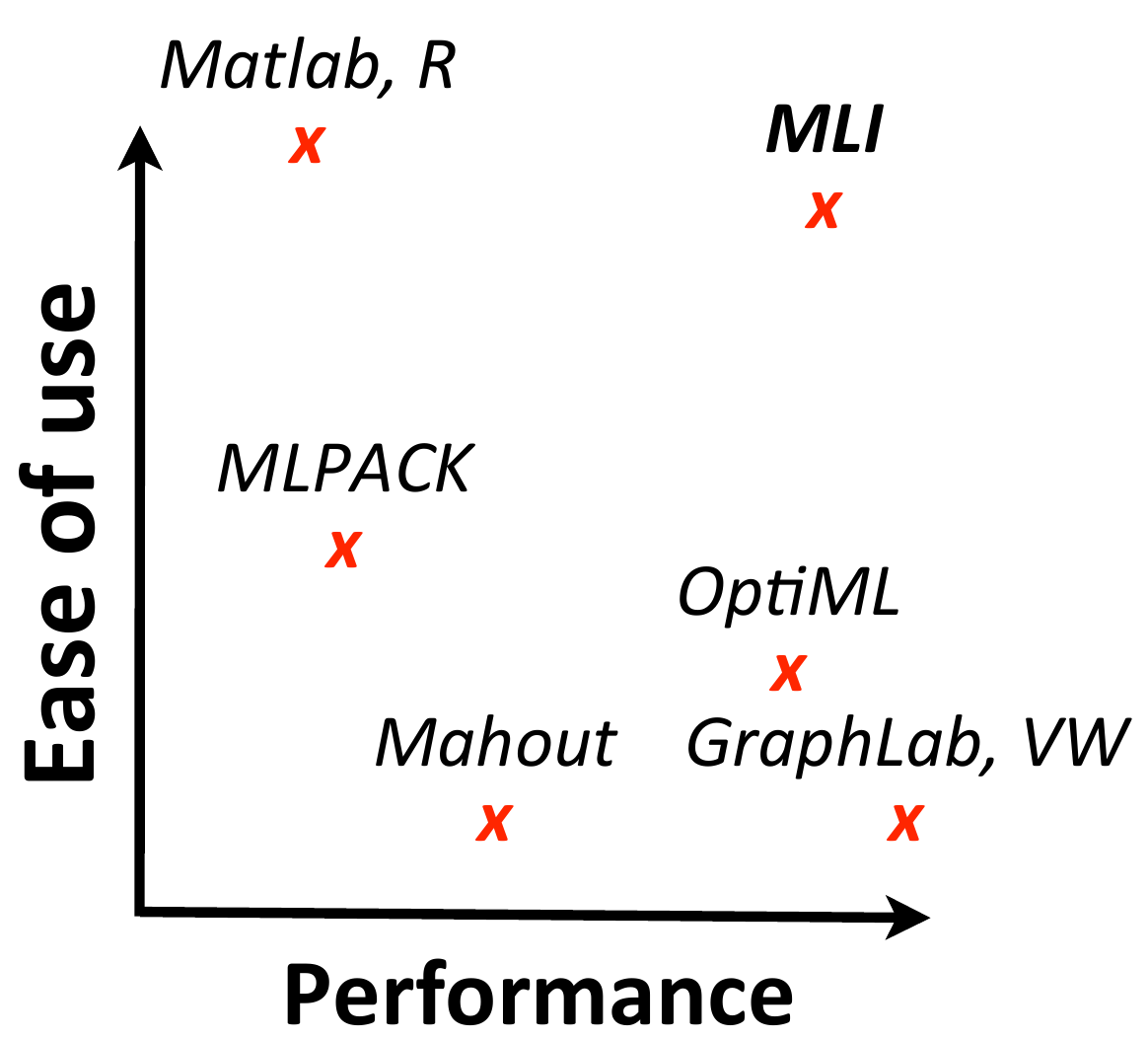}
  \caption{Landscape of existing development platforms for ML.} 
  \label{fig:lay-of-land}
\end{figure}

In this paper, we present a novel API for ML, called \mlint,\footnote{\mlint is
a component of \mlbase~\cite{Talwalkar12, Kraska13},
a system that aims to provide user-friendly distributed ML functionality for ML
experts and end users.} to bridge this gap between prototypes and
industry-grade ML software.  
We provide abstractions that simplify ML development in comparison to pure
MapReduce and graph-based primitives, while nonetheless allowing developers
control of the communication and parallelization patterns of their algorithms,
thus obviating the need for a complex optimizer. 
With \mlint, we aim to be in the top right corner of Figure~\ref{fig:lay-of-land},
by providing a development environment that is nearly on par with the usability
of MATLAB or R, while matching the scalability of and approaching the
walltime of low-level distributed systems.  We make
the following contributions in this work:

\noindent\textbf{\mlint}: We show how \mlint-supported
high-level ML abstractions naturally target common ML problems related to data
loading, feature extraction, model training and testing.


\noindent\textbf{Usability}: We demonstrate that implementing ML algorithms written 
against \mlint 
yields concise, readable code, comparable to MATLAB or R.

\noindent\textbf{Scalability}: We implement \mlint on
Spark~\cite{Spark}, a cluster computing system designed for iterative
computation in a large scale distributed setting.
Our results with logistic regression and 
matrix factorization illustrate that \mlint/Spark vastly outperforms Mahout and
matches the scaling properties of specialized, low-level systems (Vowpal
Wabbit, GraphLab), with performance within a small constant factor.

\section{Related work}

%
%

The widespread application of ML techniques has 
inspired the development of new ML platforms 
with focuses ranging from productivity to scalability.  In this section we
review a few of the representative projects.
It is worth nothing that in many cases \mlint is inspired and guided
by these earlier efforts.


Systems like MATLAB and R pioneered high-productivity numerical
computing. By combining high-level languages tailored to application
domains with intuitive interfaces and interactive interpreted
execution, these systems have redefined the way scientists and
statisticians interact with data. 
%
From the database community, projects like MADLib~\cite{MADLib} and Hazy~\cite{hazy} have
tried to expose ML algorithms in the context of well
established systems.  Alternatively, projects like Weka~\cite{weka},
scikit-learn~\cite{scikitlearn} and Google Predict~\cite{google_prediction}
have sought to expose a library of ML tools in an intuitive
interface.  
However, none of these systems focus on the challenges of scaling ML 
to the emerging distributed data setting. 



High productivity tools have struggled to keep up with the growing
computational, bandwidth, and storage demands of modern large-scale ML.
Although both MATLAB and R now provide limited support for multi-core and
distributed scaling, they adopt the traditional process centric approach and
are not well suited for large-scale distributed
data-centric workloads~\cite{presto,matlabcluster}. In the R community, efforts
have been made to run R on data-centric runtimes like Hadoop~\cite{revr},
but to our knowledge none have obtained widespread adoption.

Early efforts to develop more scalable tools and APIs for ML 
focused on specific applications.  Systems like liblinear~\cite{liblinear},
Vowpal Wabbit~\cite{vw}, and Shogun~\cite{Shogun} initially focused on linear
models, online learning, and kernel methods, respectively.  Others, like
MLPack~\cite{mlpack2011}, started to develop entire collections of learning
algorithms optimized for multicore architectures.  These efforts lead to highly
efficient systems for specialized tasks, but do not directly simplify the
design and implementation of new scalable ML methods, and most are not
well-suited to distributed learning.

Various methods have leveraged MapReduce platforms like Hadoop to
develop distributed ML libraries.
Mahout~\cite{mahout} 
does not simplify the design and development of new ML methods,
and its reliance on HDFS to store and communicate intermediate state makes it
poorly suited for iterative algorithms. 
SystemML~\cite{SystemML} introduces a low-level algebra which
it then compiles to MapReduce jobs. This algebra
exposes the opportunity for advanced optimization, but also complicates the
system, and SystemML also
suffers from Hadoop's limitations 
on iterative computation.

Others have sought to generalize the MapReduce computational model.
Systems like DryadLinq \cite{dryad} and Hyracks \cite{Hyracks} can efficiently
execute complex distributed data-flow operations and express full relational
algebras.  However, these systems expose low-level APIs and require the ML 
expert to recast their algorithms as dataflow operators.  In contrast,
GraphLab~\cite{Gonzalez12} 
is well suited to certain types of ML 
tasks, its low-level API and focus on graphs makes it challenging to
apply to more traditional ML problems.
Alternatively, OptiML \cite{optiml} and SEJITS \cite{sejits} provide higher
level embedded DSLs capable of expressing both graph and matrix operations and
compiling those operations down to hardware accelerated routines.  Both 
rely on pattern matching to find regions of code which can be
mapped to efficient low-level implementations.  Unfortunately, finding common
patterns can be challenging in rapidly evolving disciplines like machine
learning.

\section{\mlint}

With \mlint, we introduce mild constraints on the computational model in order to promote the design
and implementation of user-friendly, scalable algorithms.
Our interface consists of two fundamental objects -- MLTable and LocalMatrix -- each with its own API.
These objects are used by developers to build Optimizers, which in turn are used by Algorithms to produce Models. 
It should be noted that these APIs are system-independent - that is, they can be implemented in local and distributed settings (Shared Memory, MPI, Spark, or Hadoop). 
Our first implementation supporting this interface is built on the Spark platform.

These APIs help developers solve several common problems. First, MLTable assists in data loading and feature extraction.
While other systems~\cite{vw,mahout,Gonzalez12,Low10} require data to be imported in custom formats, MLTable allows users to load their data in an unstructured or semi-structured format, apply a series of transformations, and subsequently train a model.
LocalMatrix provides linear algebra operations on subsets of the fully featurized dataset. 
By breaking the full dataset into row-wise partitions of the original dataset
and operating locally on those partitions, we give the developer access to
higher level programming primitives while still allowing them to control the
communication that takes place between nodes and reason about the computational
complexity of their algorithms.  

As part of \mlint, we also pre-define a set of
common interfaces for Optimization, Algorithms, and Models to encourage code
reuse and to ensure a consistent external system interface. In the remainder of
this section we describe these abstractions in more detail.

\subsection{MLTable}
MLTable is an object which provides a familiar table-like interface to a
developer, and is designed to mimic a SQL table, an R \texttt{data.frame}, or a
MATLAB Dataset Array.  The basic MLTable API is illustrated in Figure~\ref{fig:mltableinterface}.  An MLTable is a collection of \emph{rows},
each of which conforms to the table's \emph{column schema}.  Each column is of
a particular type, optionally has a name, and can be of the following basic
types: String, Integer, Boolean, and Scalar (floating point numeric data).
Importantly, any cell in the table can be ``Empty'' and this is represented
with a special value.  The table interface, which should be familiar to many
developers, supports common operations like relational joins, unions, and
projections - as well as map and reduce operations on rows which follow similar
semantics to other MapReduce systems. 

Additionally, tables support batch operations on \emph{partitions} of the data,
which enable parallel data-local operation on multiple data items.  While
ML algorithms primarily expect numerical data as input, we expose
MLTable as an interface for processing the semi-structured, mixed type data
that are present in real-world applications, and transforming this raw data into feature
vectors for model training.  Given this interface, we are able to load
structured data into an MLTable, and then apply a series of transformations to
the data in parallel to produce input that is suitable for a ML
algorithm.  By supporting common data integration tasks out of the box in a
straightforward and consistent manner, we significantly decrease the amount of
time spent during data preparation and feature extraction.

Once data is featurized, it can be cast into an MLNumericTable, which is a convenience type that most ML algorithms will expect as input. 
The MLNumericTable interface is the same as MLTable, but it guarantees that all columns are numeric, and by convention each row will be treated as a single feature vector. 

An example of an end-to-end text clustering pipeline using MLTable is shown
in Figure~\ref{code:dataprep}.  This pipeline consists of \texttt{nGrams()} computation 
on the raw text input data and subsequent  \texttt{tfIdf()} feature extraction. 
We then perform
K-means clustering on the resulting features to produce an output model.  This model could be
used to make recommendations or as input to downstream analytical processing.

\subsection{LocalMatrix}
At their core, many ML algorithms are concisely expressed using linear algebra operations. 
For example, the update step in stochastic gradient descent for generalized linear models
such as logistic regression, linear regression, etc. involves computing
the gradient of a weight vector with respect to a test class and a training
point.  In the case of logistic regression, this is ultimately the dot product
of two vectors, (or a matrix/vector multiplication in the case of mini-batch
SGD), followed by a vector/vector subtraction.

LocalMatrix provides these linear algebra primitives but on \emph{partitions}
of data.  The partitions of the data presented to the developer
are typically automatically determined by the system.  That is, we require
programmers to develop algorithms such that all  operations can be
performed locally and later combined via global \texttt{reduce} operations.
This re-assembles to a large degree the shared nothing principle from
distributed computing and often leads to highly scalable algorithms.  We also
considered exposing globally distributed linear algebra operations, but
explicitly decided against it primarily because global operators would hide the
computational complexity and communication overhead of performing these
operations.
Instead, by offering linear algebra on subsets (i.e., partitions) of the data,
we provide developers with a high level of abstraction while encouraging
them to reason about efficiency. 

Aside from the semantic difference that operations are performed on individual partitions, 
LocalMatrix is designed to resemble a \texttt{matrix} in MATLAB, R, or most other numerical programming environments. 
It supports indexing by rows, columns, or slices of each. 
A LocalMatrix also supports Matrix-Matrix and Matrix-Scalar algebraic operations, and common linear algebra routines like matrix inversion.


\subsection{Optimization, Models, and Algorithms}
In addition to MLTable and LocalMatrix, we provide additional interfaces called
Optimizer, Algorithm, and Model.

Many models cannot be solved via closed form solutions, and even when
closed-form solutions exist, the computational complexity of these solutions
often increases super-linearly with data size, as in the case with basic linear
regression.  As a result, various optimization techniques are used to converge
to an approximate solution while iterating over the data.  We treat
optimization as a first class citizen in our API, and the system is built to support new optimizers.  We refer the reader to our reference implementation for Stochastic
Gradient Descent in Figure~\ref{code:lr}.

Finally, we encourage developers to implement their algorithms using the
Algorithm interface, which should return a model as specified by the Model
interface.  An algorithm implementing the Algorithm interface is a class with a
\texttt{train()} method that accepts data and hyperparameters as input, and
produces a Model.  A Model is an object which makes predictions.  In the case of
a classification model, this would be a predicted class given a new example
point.  In the case of a collaborative filtering model, this might be
recommendations for an existing user in the system.  Both interfaces are rather
simple, but crucially help to provide one common interface for developers (and to
the \mlbase system as a whole). 

\section{Examples}
To evaluate the design claims made in the earlier sections, we evaluate \mlint 
as well as competing ML systems on two representative real-world problems,
namely binary classification and matrix factorization.
When implementing algorithms against \mlint, we chose Spark as
our first platform because it is well-suited for 
computationally intensive, iterative jobs on large datasets that are
characteristic of large ML workloads.  Moreover, many large-scale ML systems,
e.g, ~\cite{vw, Gonzalez12} do not emphasize fault tolerance. In
contrast, Spark's resilience
properties, due to automatic data replication and computation lineage, are
quite attractive 
in a distributed environments where
automatic recovery from node failure is a necessity. 
Given our choice to build on top of Spark, it was natural
for the first implementation of our API to be in Scala.

Our experiments illustrate three attractive features about \mlint.
First, we
show that \mlint yields concise and readable code.  
In the
interest of space, here we compare the code length for comparable
implementations of algorithms in MATLAB and \mlint in the Appendix.
Second, we argue that \mlint supports a wide variety of
algorithms. Although we focus on two problem settings, these examples
demonstrate wide-ranging functionality of \mlint and in fact naturally extend
to a diverse group of ML algorithms, e.g., linear SVMs, linear regression, and
(L1, L2, elastic net)-regularized variants therein, simply by changing the
expression of the gradient function (and adding a proximal operator in the case
of L1-regularization).
Third, we demonstrate that the implementations written against \mlint 
are performant and scalable. We
present performance results comparing execution times of various systems on our two
examples.  We further present extensive strong and weak scalability results.
Both sets of results show that the implementations in \mlint match the
scalability of low-level distributed systems with performance within a small
constant factor.

\begin{figure*}[ht!]
\begin{center}
\begin{tabular} {@{}c@{}c@{}c@{}}
\begin{minipage}[c]{.3\textwidth}
   \begin{tabular}{| c | c |}
    \hline
        \textbf{System} & \textbf{Lines of Code}\\ \hline
    \mlint & 55 \\ \hline
    Vowpal Wabbit & 721 \\ \hline
    MATLAB & 11 \\ \hline
    \end{tabular}
\end{minipage} &
\begin{minipage}[c]{.34\textwidth}
\ipsfig{.24}{figure=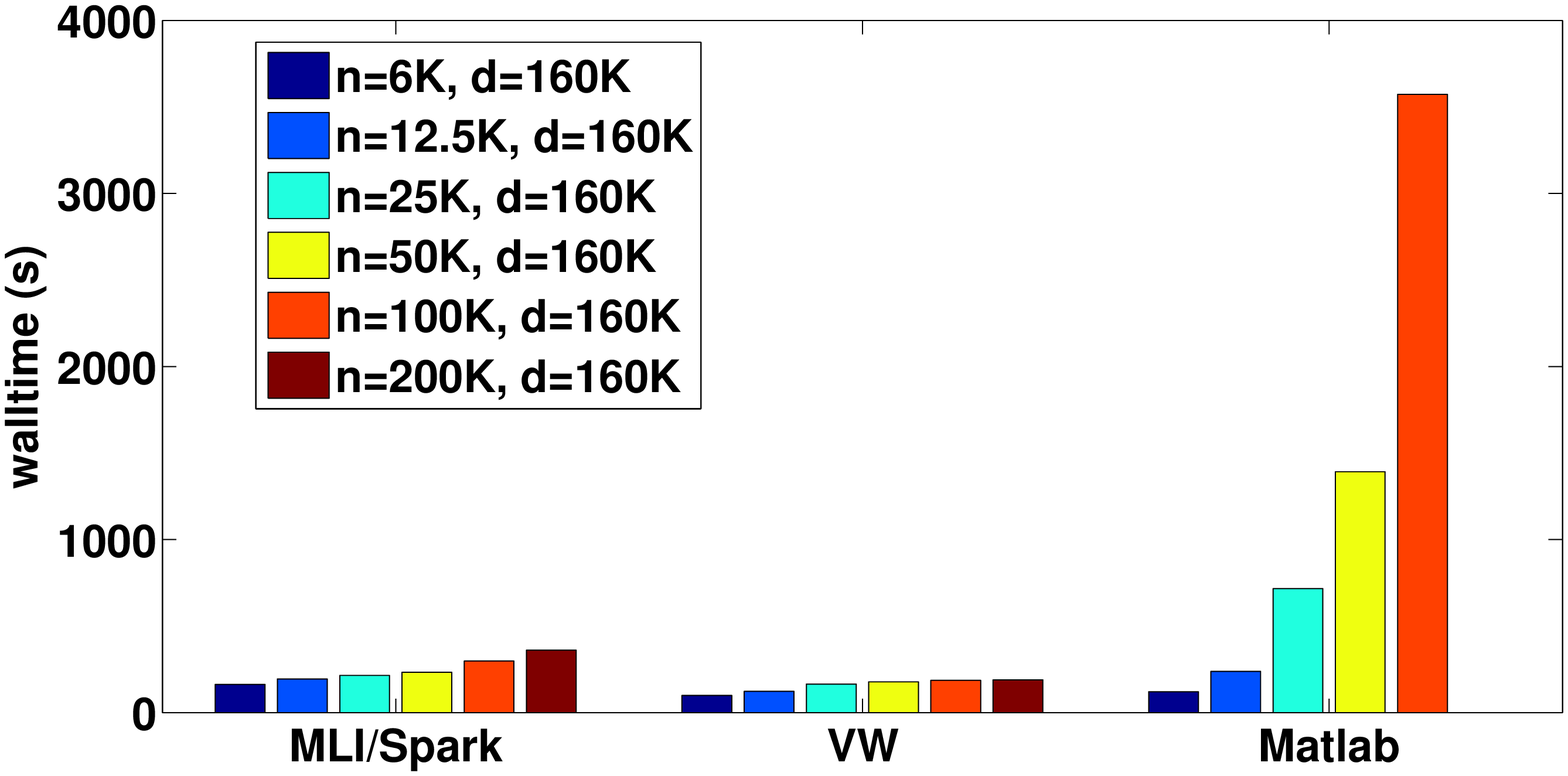}
\end{minipage} &
\qquad \begin{minipage}[c]{.32\textwidth}
\ipsfig{.30}{figure=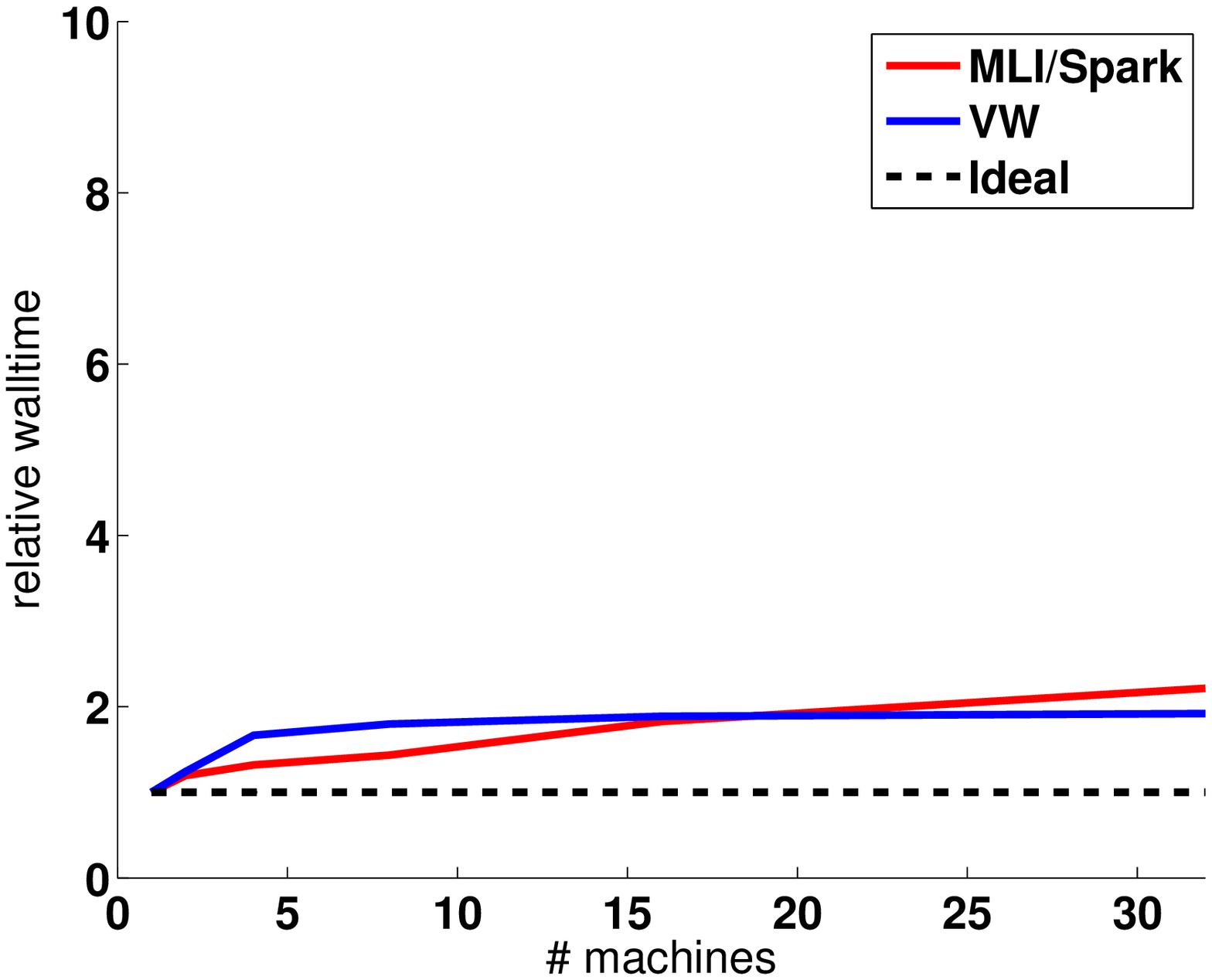} 
\end{minipage} \\ 
(a) & (b) & (c) \\ 
\end{tabular}
\end{center}
\vspace{-2mm}
\caption{Logistic regression experiments. (a) Lines of code. (b)
Execution time for weak scaling. (c) Weak scaling.}
\vspace{-5mm}
\label{fig:logreg_experiments}
\end{figure*} 

\ignore{
\begin{figure*}[ht!]
\begin{minipage}[l]{1.0\columnwidth}
\ipsfig{.24}{figure=results/performance_logreg.eps}
\caption{Execution time for weak scaling for logistic regression.}
\label{fig:lr_weak_performance}
\end{minipage}
\begin{minipage}[l]{1.0\columnwidth}
\begin{center}
  \ipsfig{.30}{figure=results/weak_scaling_logreg.eps}
\end{center}
\caption{Weak scaling for logistic regression.}
\label{fig:lr_weak_scaling}
\end{minipage}
\end{figure*}

\begin{table}
\centering
  \normalsize
   \begin{tabular}{| c | c |}
    \hline
        \textbf{System} & \textbf{Lines of Code}\\ \hline
    \mlint & 60 \\ \hline
    Vowpal Wabbit & 721 \\ \hline
    MATLAB & 11 \\ \hline
    \end{tabular}
  \caption{Lines of code for logistic regression \vspace{-1em}}
   \label{tab:lr_lines}
\end{table}
}

\subsection{Binary Classification: Logistic Regression}
Let $X \in \reals^{n \times d}$ be a dataset of $n$ points with $d$ features,
$x_i \in \reals^{d}$ be the $i$th data point, and define $y \in \{0,1\}^n$ as
the corresponding set of binary labels. Logistic regression is a canonical
classification algorithm. 
The optimal parameter vector $w^*\in\reals^d$ can be found by
minimizing the negative likelihood function, $f(w) = -\log p(X|w)$.
Taking
the gradient of the negative log likelihood, we have:
\begin{equation}
\label{eq:grad}
\nabla f = \sum_{i=1}^n \bigg [\big (\sigma(w^\top x_i)
- y_i \big ) x_i \bigg ]\,,
\end{equation}
where $\sigma(x) = 1 / (1+\exp{-x})$ is the logistic sigmoid function. 
Gradient descent (GD) is a standard first-order iterative method to solve for
$w^*$; at the $t$th iteration we move in the direction of the negative gradient
with step size controlled by a learning rate, $\eta$, i.e., we set $w_{t+1} =
w_t - \eta \nabla f$. Stochastic gradient descent (SGD)
involves approximating the sum in Equation~\ref{eq:grad} by a
single summand.  



\subsubsection*{Experimental Setup and Data}
We run both strong and weak scaling experiments on 1, 2, 4, 8, 16, and 32
machines.  All are Amazon \texttt{m2.4xlarge} EC2 instances with 68GB of RAM
and 8 virtual cores running in the \texttt{us1-east} region.  They are
configured using the default Spark 0.7.0 AMI and are running a recent version 
of Spark and Hadoop 1.0.4.  We compare our system to the latest version of
Vowpal Wabbit (VW) running on the same cluster, and MATLAB running on a
similarly configured (single node) machine.  We do not compare
against Mahout for these experiments because its implementation of Logistic
Regression via SGD is very communication intensive, and we feel that
this implementation would not provide a fair comparison for Mahout.

We run our weak scaling experiments on a training set of up to approximately
200GB of featurized ImageNet~\cite{imagenet_cvpr09} data where each image is represented with 160K
dense features, yielding approximately 200K images total for the 32-node
experiment.  The number of input points used is proportional to the number of
nodes in the cluster for the experiment.  We further note that this experiment
only represents approximately 20\% of the full ImageNet dataset.  While we were
able to train a full classifier using our system in approximately 2.5 hours,
the preprocessing required to prepare the data for VW on the full set of data
was too onerous to complete the experiment.  In our strong scaling experiments,
we train on 5\% of this base data for the same number of nodes.

\subsubsection*{Implementation}
We have implemented logistic regression via SGD. To approximate the algorithm
used in VW~\cite{vwsgd} we run SGD locally on each partition
before averaging parameters globally.  We note, however, that there are several
alternative methods to implement SGD on top of \mlint.  Implementing Logistic Regression in \mlint is as simple as defining the form of the gradient function and calling the SGD Optimizer with that function.  Additionally, the code that implements
\emph{StochasticGradientDescent} is both short and
fairly interpretable.

Algorithmically, our implementation is identical to VW, with one meaningful
difference, namely aggregating results across worker nodes after each round.
VW uses an ``AllReduce'' communication primitive to build an aggregation tree
when averaging together model parameters after each iteration.  It then uses
the same tree to broadcast these results back to workers.  In contrast, we take
a more traditional MapReduce approach and average all parameters at the
cluster's master node at each iteration, then broadcast the parameters to each
node using a one-to-many broadcast.  As the number of machines increases, VW's
approach is theoretically more efficient from the perspective of communication
and parallelizes better with respect to computation.  In practice, we see
comparable scaling results as more machines are added.

In MATLAB, we implement gradient descent instead of SGD, as gradient descent
requires roughly the same number of numeric operations as SGD but does not
require an inner loop to pass over the data. It can thus be implemented in a
`vectorized' fashion, which leads to a significantly more favorable runtime.
We show MATLAB's performance here as a reference for
training a model on a similarly sized dataset on a single multicore machine.

\subsubsection*{Results}

In our weak scaling experiments (Figures~\ref{fig:logreg_experiments}b, \ref{fig:logreg_experiments}c),
we can see that our clustered system begins to
outperform MATLAB at even moderate levels of data, and while MATLAB runs out of
memory and cannot complete the experiment on the 200K point dataset, our system
finishes in less than 10 minutes. Moreover, the highly specialized VW is on
average 35\% faster than our system, and never twice as fast.  These times
\emph{do not} include time spent preparing data for input input for VW, which
was significant, but we expect that these would be a one-time cost in a production
environment.

From the perspective of strong scaling our solution which is presented in the Appendix actually outperforms VW in raw
time to train a model on a fixed dataset size when using 16 and 32 machines,
and exhibits better strong scaling properties, much closer to the gold standard of
linear scaling for these algorithms. We are unsure whether this is due to our
simpler (broadcast/gather) communication paradigm, or some other property of
the system.

\begin{figure*}[ht!]
\begin{center}
\begin{tabular} {@{}c@{}c@{}c@{}}
\begin{minipage}[c]{.3\textwidth}
\begin{tabular}{| c | c |}
    \hline
        \textbf{System} & \textbf{Lines of Code}\\ \hline
    \mlint & 35 \\ \hline
    GraphLab & 383 \\ \hline
    Mahout & 865 \\ \hline
    MATLAB-Mex & 124 \\ \hline
    MATLAB & 20 \\ \hline
    \end{tabular} 
\end{minipage} &
\begin{minipage}[c]{.37\textwidth}
\ipsfig{.26}{figure=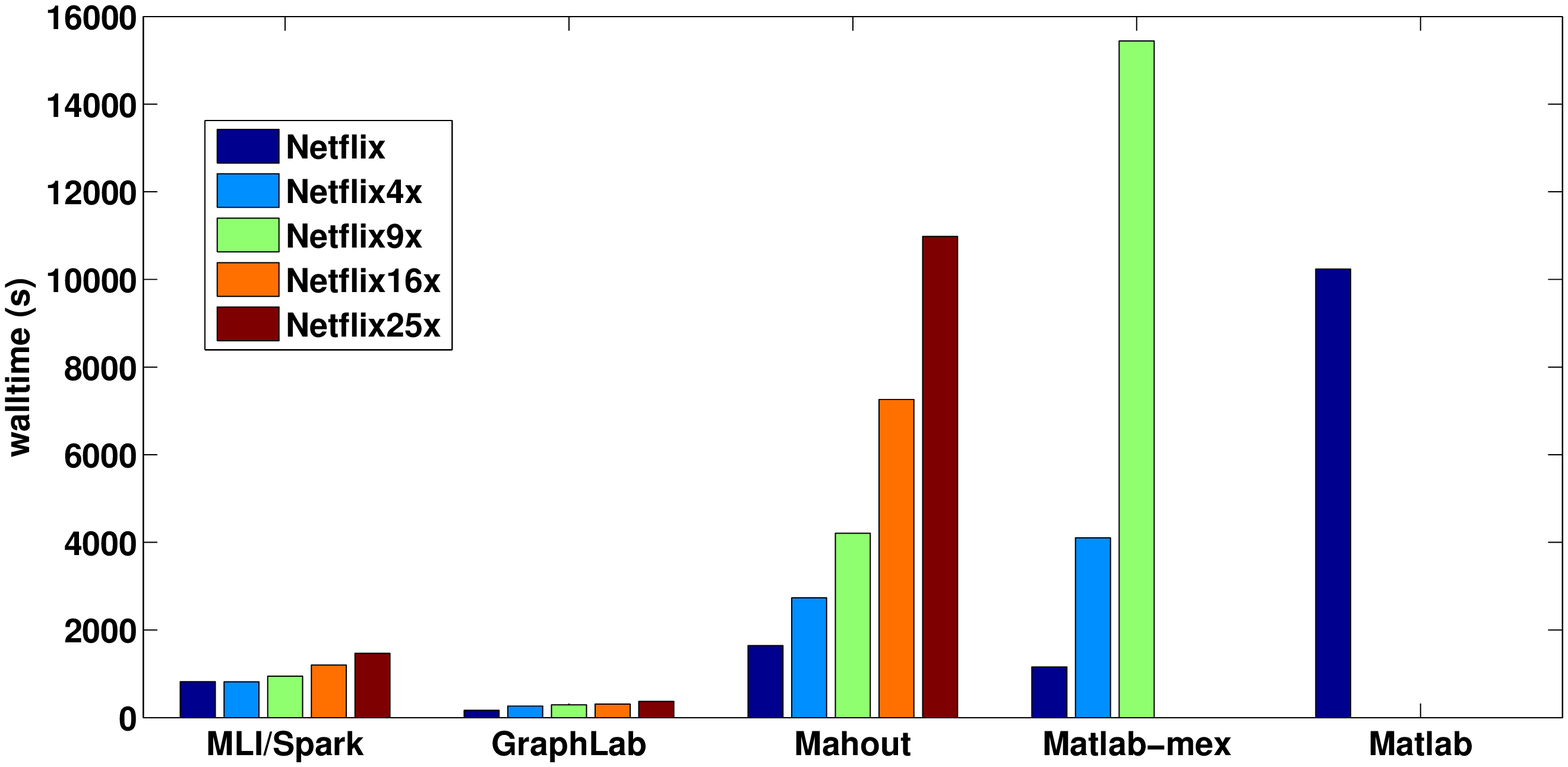}
\end{minipage} &
\qquad \begin{minipage}[c]{.32\textwidth}
\ipsfig{.30}{figure=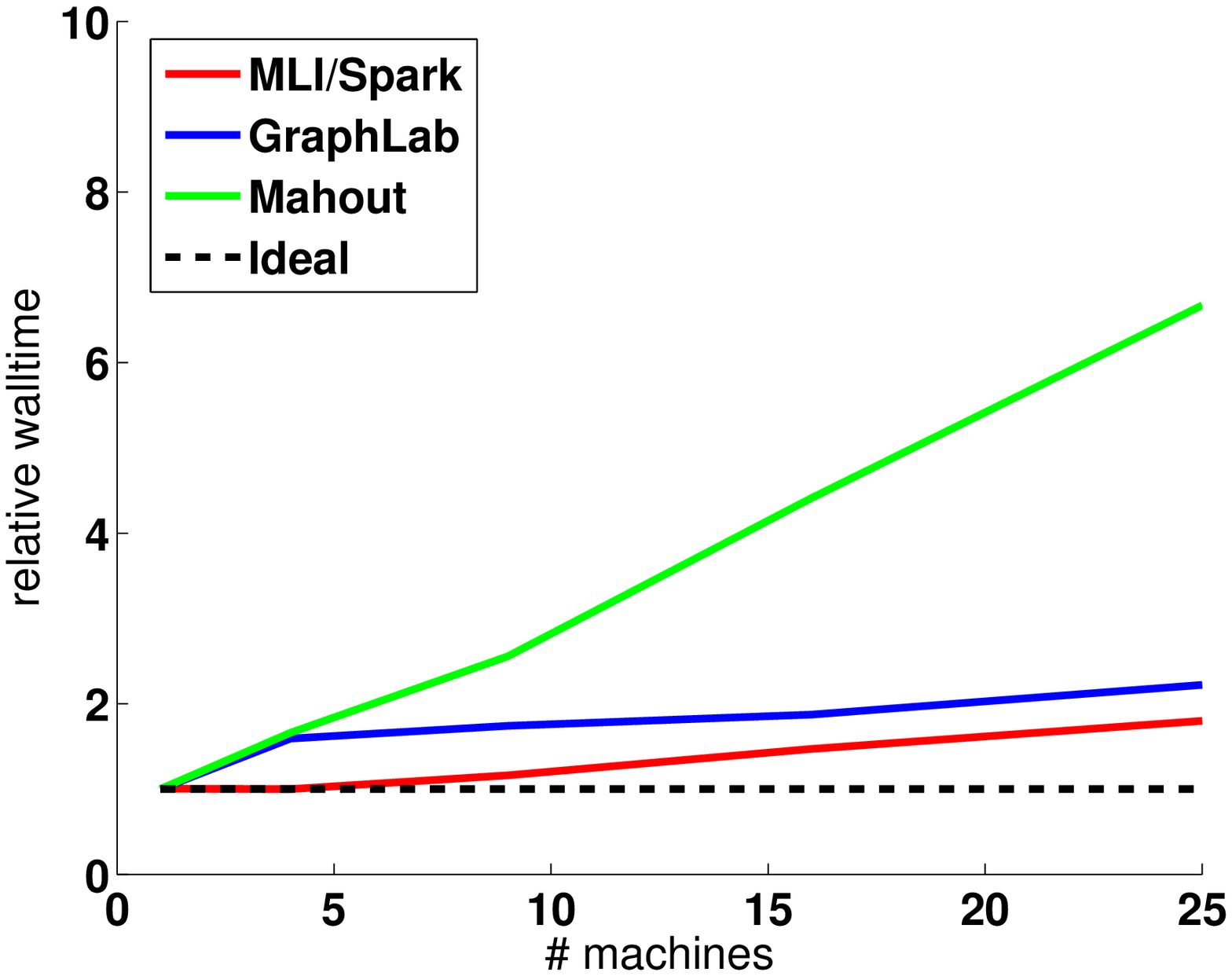}
\end{minipage} \\ 
(a) & (b) & (c) \\ 
\end{tabular}
\end{center}
\vspace{-2mm}
\caption{ALS experiments. (a) Lines of code. (b)
Execution time for weak scaling. (c) Weak scaling. }
\vspace{-5mm}
\label{fig:als_experiments}
\end{figure*} 

\ignore{
\begin{figure*}[ht!]
\begin{minipage}[l]{1.0\columnwidth}
   \begin{tabular}{| c | c |}
    \hline
        \textbf{System} & \textbf{Lines of Code}\\ \hline
    \mlint & 32 \\ \hline
    GraphLab & 383 \\ \hline
    Mahout & 865 \\ \hline
    MATLAB-Mex & 124 \\ \hline
    MATLAB & 20 \\ \hline
    \end{tabular}
   \label{tab:als_lines}
  \caption{Lines of code for various implementations of ALS \vspace{-1em}}
\end{minipage}

\begin{minipage}[l]{1.0\columnwidth}
\ipsfig{.30}{figure=results/performance_als.eps}
\caption{Execution time for weak scaling for ALS.}
\label{fig:als_weak_performance}
\end{minipage}
\begin{minipage}[l]{1.0\columnwidth}
\begin{center}
  \ipsfig{.30}{figure=results/weak_scaling_als.eps}
\end{center}
\caption{Weak scaling for ALS.}
\label{fig:als_weak_scaling}
\end{minipage}
\end{figure*}

\begin{table}
\centering
  \normalsize
   \begin{tabular}{| c | c |}
    \hline
        \textbf{System} & \textbf{Lines of Code}\\ \hline
    \mlint & 35 \\ \hline
    GraphLab & 383 \\ \hline
    Mahout & 865 \\ \hline
    MATLAB-Mex & 124 \\ \hline
    MATLAB & 20 \\ \hline
    \end{tabular}
  \caption{Lines of code for various implementations of ALS \vspace{-1em}}
   \label{tab:als_lines}
\end{table}
}

\subsection{Collaborative Filtering: Alternating Least Squares}
Matrix factorization is a technique used in recommender systems to predict
user-product associations. Let $M \in \mathbb{R}^{m \times n}$ be some
underlying matrix and suppose that only a small subset, $\Omega(M)$, of its
entries are revealed. The goal of matrix factorization is to find low-rank
matrices  $U \in \mathbb{R}^{m \times k}$ and $V \in \mathbb{R}^{n \times k}$,
where $k \ll n,m$, such that $M \approx UV^T$.
Commonly, $U$ and $V$ are estimated using the following bi-convex objective:
\begin{equation}
\label{matfact_objfn}
\min_{U,V} \sum_{(i,j) \in \Omega(M)}(M_{ij}-U_i^TV_j)^2 + \lambda(||U||_F^2+||V||_F^2) \,.
\end{equation}
Alternating least squares (ALS) is a widely used method for matrix
factorization that solves (\ref{matfact_objfn}) by alternating between
optimizing $U$ with $V$ fixed, and $V$ with $U$ fixed, using a well-known closed-form solution at each step~\cite{koren09}.

\subsubsection*{Experimental Setup and Data}
We test both strong and weak scaling experiments using 1, 4, 9, 16, and 25
machines with the same specifications as in our previous experiments.
We run our weak scaling experiments on a training set of up to approximately 50
GB of collaborative filtering data. This data is created by repeatedly tiling
the Netflix collaborative filtering dataset. This allows us to maintain the
sparsity structure of the dataset, and increase the number of parameters in a
fixed manner. For weak scaling, the size of the dataset is proportional to the
number of machines used in the cluster for the experiment. Thus, when running
the largest experiment on 25 machines, we use a dataset that is 25x the size of
the Netflix dataset.  In our strong scaling experiments, we trained on 9x the
Netflix dataset, changing only the number of machines.

For both strong and weak scaling experiments, we keep the following parameters
fixed. We run ALS for 10 iterations, use a rank of 10, and set
$\lambda = .01$. We do not calculate training
error or testing error for timing purposes, but note that ALS methods from all
systems achieved comparable error rates at the end of 10 iterations.

\subsubsection*{Implementation}
We implement ALS by updating the rows of $U$ or $V$ in parallel across
machines, and then broadcasting the factors to each machine after each update.
We distribute both the matrix $M$ and a transposed version of this matrix
across machines in order to quickly access relevant ratings.
Our reference implementation makes use of several features of \mlint, including
support for CSR-compressed sparse representations of matrices, several linear algebra
primitives, and heavy use of MLTable functionality.
Linear algebra methods such as
 matrix transpose, matrix multiplication, and solving linear systems are
 supported. LocalMatrix also supports important access methods, such as the
 \textit{nonZeroIndices}, which returns the nonzero column indices
 for a given row. 

We compare our system to the latest versions of Mahout and GraphLab on the same
cluster, and MATLAB running on a similarly configured machine. In addition, we
test a version of ALS in MATLAB using \textit{mex}, an interface that allows MATLAB to call directly into C++/Fortran routines. Comparing \mlint to these other implementations, we see that the
MLNumericTable and LocalMatrix objects provide convenient abstractions for
patterns, thus resulting in concise code. Indeed, Figure \ref{code:bals} shows
that our implementation is about the same length as the MATLAB code, while
Figure \ref{fig:als_experiments}(a) shows the stark comparison in code length in
comparison to Mahout and GraphLab. 

\subsubsection*{Results}

In our weak scaling experiments for ALS (Figures \ref{fig:als_experiments}b, \ref{fig:als_experiments}c),
we see that our system outperforms MATLAB and
the highly-optimized MATLAB-Mex, at even moderate levels of data. Both MATLAB
and MATLAB-Mex run out of memory before successfully running the 16x or 25x
Netflix datasets.  We remain within 4x of the highly specialized system
GraphLab, and maintain a similar scaling pattern.  We outperform Mahout both in
terms of total execution time for each run and scaling across cluster size. 

We see similarly promising results with our strong scaling experiments as illustrated
in 
the Appendix - with MATLAB running out of memory before completing on the
9x Netflix dataset, and GraphLab outperforming \mlint by less than a factor of
4x.

\subsection{Configuration Considerations}
Although we ran all of our experiments on comparable or identical hardware,
different software systems varied drastically in terms of ease of
installation, configuration, and executing code.

\subsubsection*{Vowpal Wabbit}
To use VW in cluster mode, users must carefully
partition their datasets into equally sized compressed files of training data,
where the total number of files should equal the number of map tasks that the
user desires to use concurrently on the cluster.  Although VW uses Hadoop Streaming to launch
cluster tasks, it eschews the traditional MapReduce paradigm in favor of AllReduce. To
support this new communication primitive, it must open a side-channel TCP
socket between map tasks to communicate incremental results. The combination leads to a 
failure-prone system as well as difficulty in data preparation.

\subsubsection*{Mahout}
Mahout is fairly easy to set up on an already existing Hadoop cluster, and
its input file formats are reasonably close to our own.  However, in order to
run Mahout effectively on problems larger than the traditional Netflix dataset,
the user must take great care to tune job memory configuration parameters
correctly to ensure that jobs complete in a performant manner. 

\subsubsection*{GraphLab}
While GraphLab performed very well in our speed and scalability tests, it was
rather difficult to set up and integrate with an existing cluster with
distributed data stored in HDFS.  In order to set up GraphLab, users must
configure their clusters with MPI, download, build and install GraphLab and its
required dependencies, and manually copy the software to each machine on the
cluster.  If a single input matrix will not fit into memory, it must be stored and
loaded as multiple separate files.  This complicates preprocessing and requires
developers to take extra steps depending on their problem sizes. 

\subsubsection*{\mlint and Spark}
Setting up and configuring our system built on \mlint and Spark is
comparatively easy.  Launching a well configured cluster required a single command, and
our software ships with all its dependencies listed in SBT, and can be compiled
and run on a cluster simply by setting a few environment variables and running
one Scala program.  New algorithms can be easily added to the system as new
Scala classes, and driver programs are easily generated based on examples in
the existing library.

\section{Conclusion}
We have presented \mlint, an API for building scalable distributed machine
learning algorithms.  We have shown that its components, MLTable and
LocalMatrix, are useful primitives for data loading and transformation as well
as data-local linear algebra operations. We have shown how these primitives can
be used to code two fairly different but representative algorithms. We
evaluated these algorithms in terms of both ease-of-development and computational performance, based on
an implementation of \mlint against Spark, comparing our system with several
existing ones.  Our results show that we can provide ML developers the tools to construct high performance distributed ML algorithms without onerous programming complexity. \mlint is a foundational layer in our larger efforts with \mlbase, a system designed to simplify large-scale ML.

\section*{Acknowledgment}
This research is supported in part by NSF award No. 1122732 (AT), NSF CISE Expeditions award CCF-1139158 and DARPA XData Award FA8750-12-2-0331, and  gifts from Amazon Web Services, Google, SAP,  Cisco, Clearstory Data, Cloudera, Ericsson, Facebook, FitWave, General Electric, Hortonworks, Huawei, Intel, Microsoft, NetApp, Oracle, Samsung, Splunk, VMware, WANdisco and Yahoo!.

\bibliographystyle{IEEEtran}
{\small{\bibliography{IEEEabrv,refs}}}
\clearpage
\appendix
\renewcommand{\thesection}{\Alph{section}}
\setcounter{figure}{0}
\setcounter{table}{0}
\makeatletter 
\renewcommand{\thetable}{A\@arabic\c@table}
\renewcommand{\thefigure}{A\@arabic\c@figure}

\noindent\begin{minipage}{\textwidth}
  \begin{tabular}{| l | l | l | p{7cm} |}
      \hline 
      \textbf{Operation} & \textbf{Arguments} & \textbf{Returns} & \textbf{Semantics} \\ \hline
      project & Seq[Index] & MLTable & Select a subset of columns from a table. \\ \hline
      union & MLTable & MLTable & Concatenate two tables with identical schemas. \\ \hline
      filter & MLRow $\Rightarrow$ Bool & MLTable & Select a subset of rows from a data table given a functional predicate.\\ \hline
      join & MLTable, Seq[Index] & MLTable & Inner join of two tables based on a sequence of shared columns.\\ \hline
      map & MLRow $\Rightarrow$ MLRow & MLTable & Apply a function to each row in the table.\\ \hline
      flatMap & MLRow $\Rightarrow$ TraversableOnce[MLRow] & MLTable & Apply a function to each row, producing 0 or more rows as output.\\ \hline
      reduce & Seq[MLRow] $\Rightarrow$ MLRow & MLTable & Combine all rows in a table using an associative, commutative reduce function.\\ \hline
      reduceByKey & Int, Seq[MLRow] $\Rightarrow$ MLRow & MLTable & Combine all rows in a table using an associative, commutative reduce function on a key-by-key basis where a key column is the first argument.\\ \hline
      matrixBatchMap & LocalMatrix $\Rightarrow$ LocalMatrix & MLNumericTable & Execute a batch function on a local partition of the data. Output matrices are concatenated to form a new table.\\ \hline
      numRows & None & Long & Returns number of rows in the table.\\ \hline
      numCols & None & Long & Returns the number of columns in the table.\\ \hline
  \end{tabular}
  \captionof{figure}{MLTable API Illustration. This table captures core operations of the API and is not exhaustive.}
  \label{fig:mltableinterface}
\end{minipage}

\noindent\begin{minipage}{\textwidth}
\centering
\begin{lstlisting}[language=Scala, firstline=6, lastline=17]
  def main(args: Array[String]) {
    val mc = new MLContext("local")

    //Read in table from file on HDFS.
    val rawTextTable = mc.textFile(args(0))
    
    //Run feature extraction on the raw text - get the top 30000 bigrams.
    val featurizedTable = tfIdf(nGrams(rawTextTable, n=2, top=30000))    
    
    //Cluster the data using K-Means.
    val kMeansModel = KMeans(featurizedTable, k=50)
  }
\end{lstlisting}
\captionof{figure}{Loading, featurizing, and learning clusters on a corpus of Text Data.}
\label{code:dataprep}
\end{minipage}

\noindent\begin{minipage}{\textwidth}
\small
\centering
  \begin{tabular}{| l | p{6cm} | l | p{5cm} |}
      \hline 
      \textbf{Family} & \textbf{Example Uses} & \textbf{Returns} & \textbf{Semantics} \\ \hline
      Shape & dims(mat), mat.numRows, mat.numCols & Int or (Int,Int) & Matrix dimensions.\\ \hline
      Composition & matA on matB, matA then matB & Matrix & Combine two matrices row-wise or column-wise.\\ \hline
      Indexing & mat(0,??), mat(10,10), mat(Seq(2,4), 1) & Matrix or Scalar & Select elements or sub-matrices.\\ \hline
      Reverse Indexing & mat(0,??).nonZeroIndices & Seq[Index] & Find indices of non-zero elements.\\ \hline
      Updating & mat(1,2) = 5, mat(1, Seq(3,10)) = matB & None & Assign values to elements or sub-matrices.\\ \hline
      Arithmetic & matA + matB, matA - 5, matA / matB & Matrix & Element-wise arithmetic between matrices and scalars or matrices.\\ \hline
      Linear Algebra & matA times matB, matA dot matB, matA.transpose, matA.solve(v), matA.svd, matA.eigen, matA.rank & Matrix or Scalar & Basic and extended linear algebra support. \\ \hline
  \end{tabular}
  \captionof{figure}{LocalMatrix API Illustration}
  \label{fig:mlmatrixinterface}
\end{minipage}

\begin{figure*}
\begin{lstlisting}[language=MATLAB, firstline=1]
function w = log_reg(X, y, maxiter, learning_rate)
  [n, d] = size(X);
  w = zeros(d,1);
  for iter = 1:maxiter
    grad = X' * (sigmoid(X * w) - y);           
    w = w - learning_rate * grad;
  end
end

% applies sigmoid function component-wise on the vector x 
function s = sigmoid(x)
  s = 1 ./ (1 + exp(-1 .* x));
end
\end{lstlisting}
\begin{lstlisting}[language=Scala, firstline=20,label=code:lr]
object LogisticRegressionAlgorithm extends NumericAlgorithm[LogisticRegressionParameters] {

  def defaultParameters() = LogisticRegressionParameters()

  def sigmoid(z: Double): Double = 1.0/(1.0 + math.exp(-1.0*z))

  def train(data: MLNumericTable, params: LogisticRegressionParameters): LogisticRegressionModel = {
    val d = data.numCols-1

    def gradient(vec: MLVector, w: MLVector): MLVector = {
      val x = MLVector(vec.slice(1,vec.length))
      x times (sigmoid(x dot w) - vec(0))
    }

    //Run gradient descent on the data.
    val optParams = StochasticGradientDescentParameters(
      wInit = MLVector.zeros(d),
      grad = gradient,
      learningRate = params.learningRate)
    val weights =StochasticGradientDescent(data, optParams)

    new LogisticRegressionModel(data.toMLTable, params, weights)
  }
}

\end{lstlisting}
\begin{lstlisting}[language=Scala, firstline=5]

object StochasticGradientDescent extends MLOpt with Serializable {

  def apply(data: MLNumericTable, params: StochasticGradientDescentParameters): MLVector = {
    var weights = wInit
    val n = data.numRows
    var i = 0

    //Main loop of SGD. Calls local SGD and averages parameters.
    while(i < params.maxIter) {
      weights = data.matrixBatchMap(localSGD(_, weights, params.learningRate, params.grad)).reduce(_ plus _) over data.partitions.length
      i+=1
    }
    weights
  }

  def localSGD(data: LocalMatrix, weights: MLVector, lambda: Double, gradientFunction: (MLVector, MLVector) => MLVector): LocalMatrix = {
    var localWeights = weights
    for (i <- data.toMLVectors) {
      //Compute the gradient and update the model.
      val grad = gradientFunction(i, loc)
      localWeights = localWeights minus (grad times lambda)
    }
    localWeights
  }
}

\end{lstlisting}
\caption{Logistic Regression Code in MATLAB (top) and \mlint (middle, bottom).}
\label{code:lr}
\end{figure*}

\begin{figure*}[ht!]
\begin{minipage}[r]{1.0\columnwidth}
\ipsfig{.27}{figure=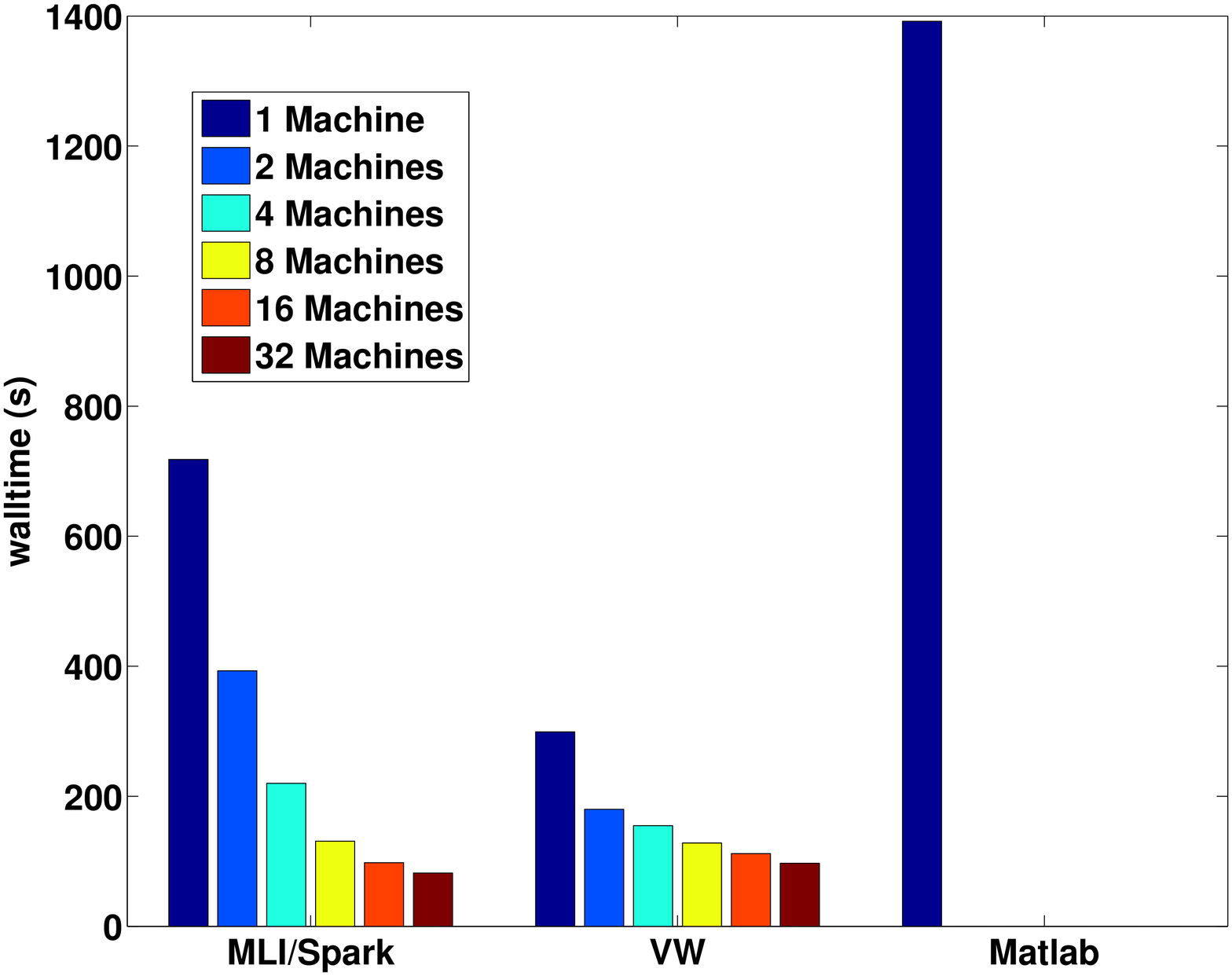}
\caption{Execution time for strong scaling for logistic regression.}
\label{fig:lr_strong_performance}
\end{minipage}
\begin{minipage}[l]{1.0\columnwidth}
\begin{center}
\hspace{3em}
  \ipsfig{.30}{figure=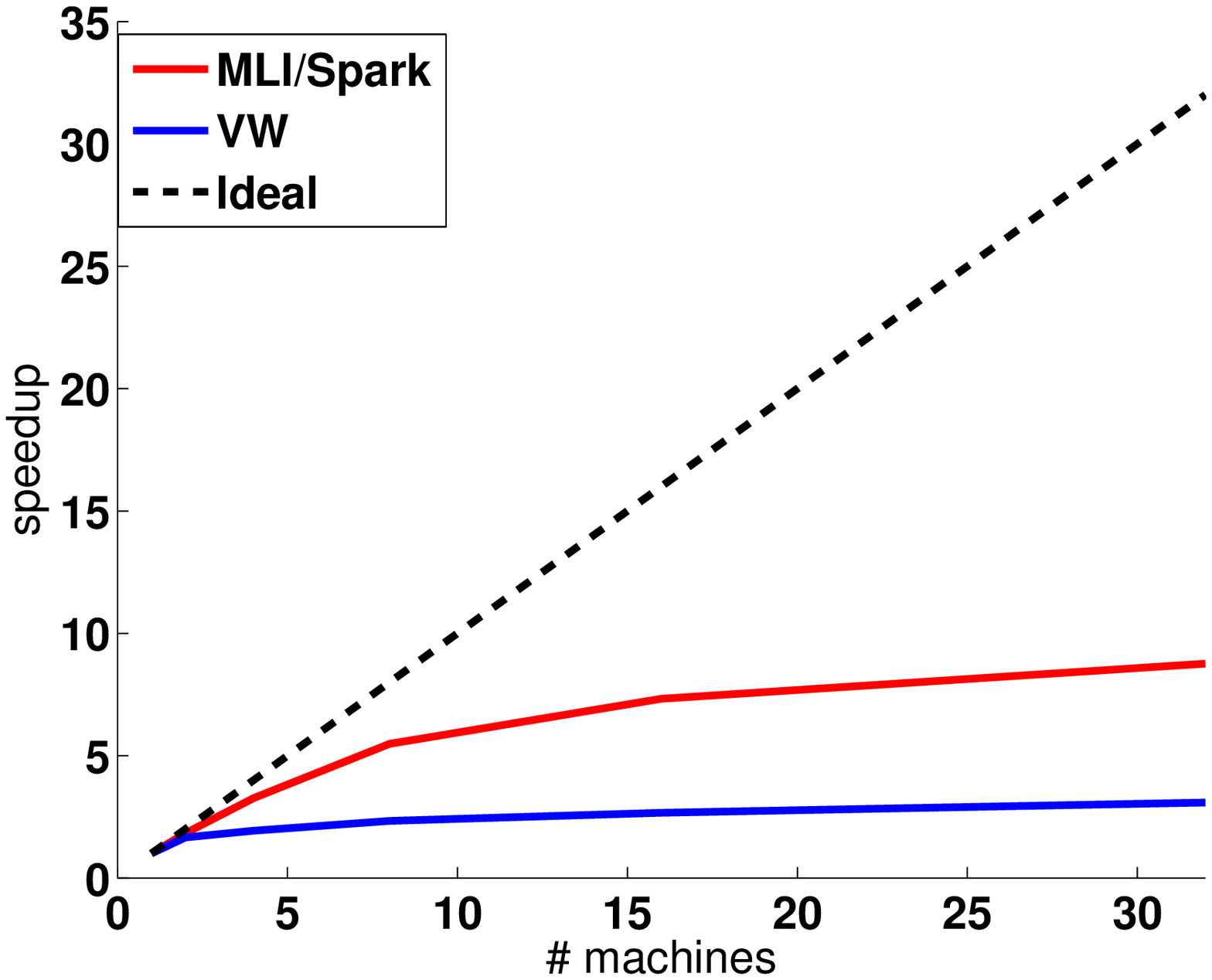}
\end{center}
\caption{Strong scaling for logistic regression}
\label{fig:lr_strong_scaling}
\end{minipage}
\end{figure*}

\begin{figure*}[ht!]
\begin{minipage}[r]{1.0\columnwidth}
\ipsfig{.27}{figure=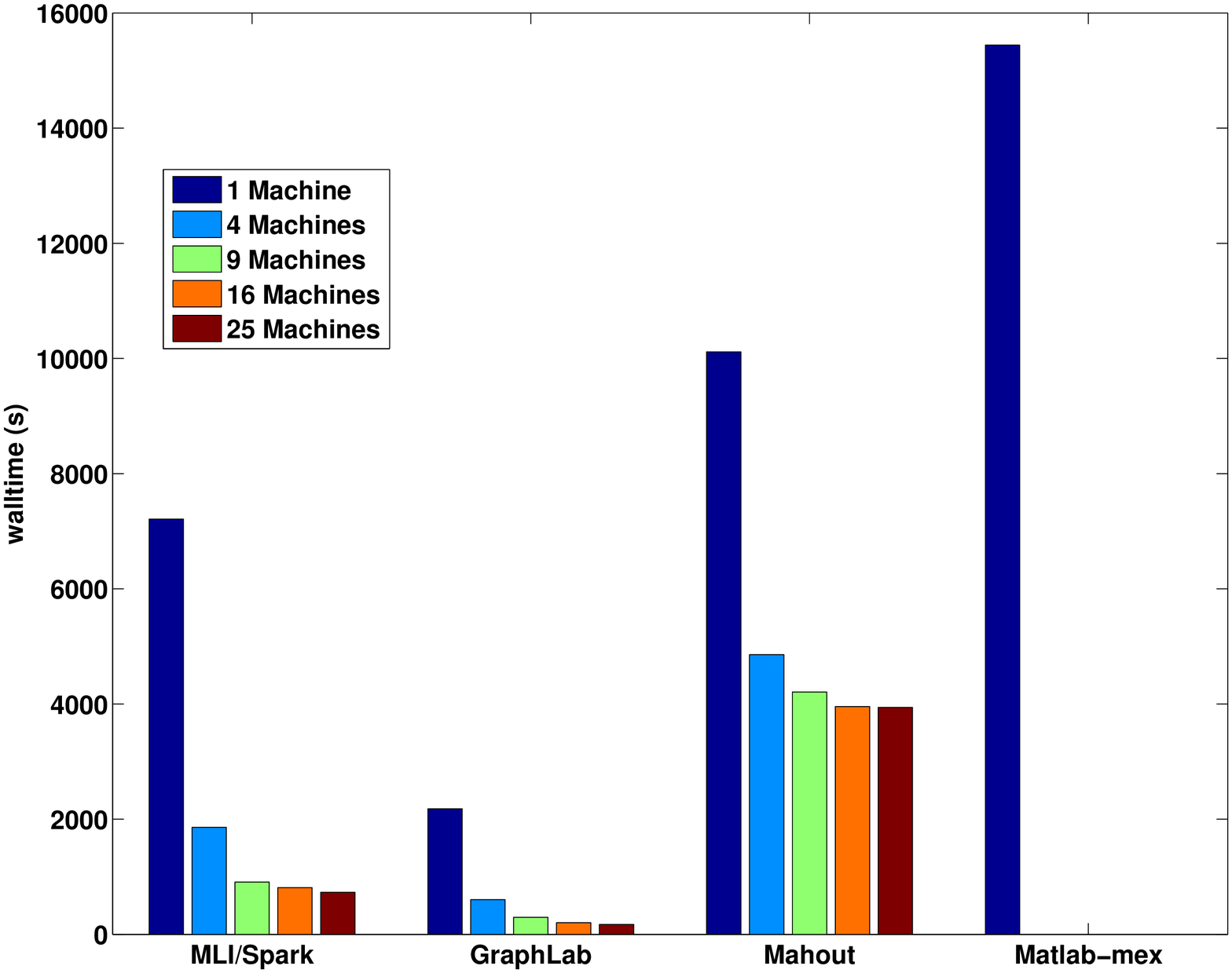}
\caption{Execution time for strong scaling for ALS.}
\label{fig:als_strong_performance}
\end{minipage}
\hspace{1em}
\begin{minipage}[l]{1.0\columnwidth}
\vspace{1em}
\begin{center}
  \ipsfig{.30}{figure=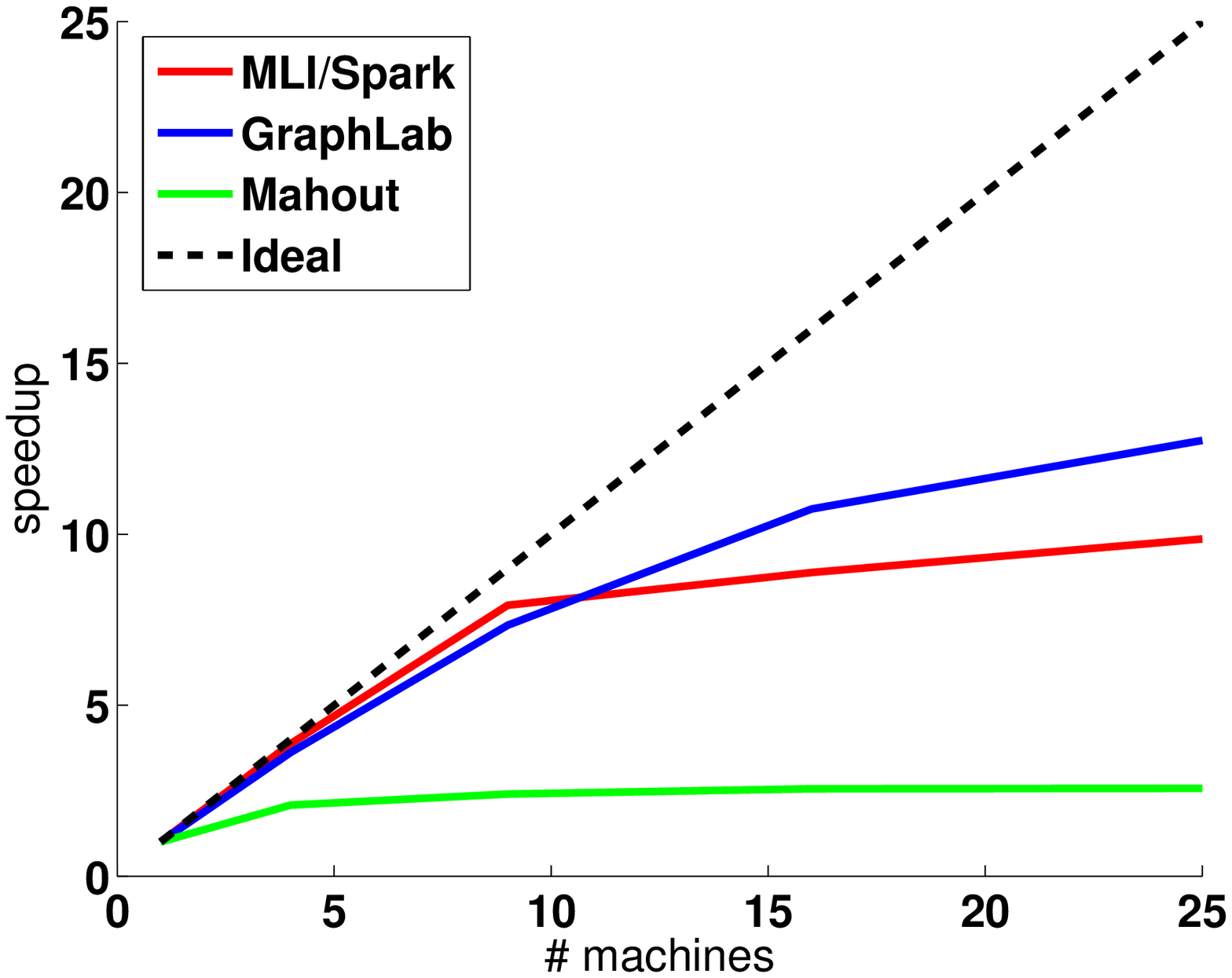}
\end{center}
\caption{Strong scaling for ALS}
\label{fig:als_strong_scaling}
\end{minipage}
\end{figure*}

\begin{figure*}
\begin{lstlisting}[language=MATLAB, firstline=22]
function [U, V] = ALS_matlab(M, U, V, k, lambda, maxiter)
 
  % Initialize variables
  [m,n] = size(M);
  lambI = lambda * eye(k);
  for q = 1:m
    Vinds{q} = find(M(q,:) ~= 0); 
  end
  for q=1:n
    Uinds{q} = find(M(:,q) ~= 0); 
  end

  % ALS main loop
  for iter=1:maxiter
    parfor q=1:m
      Vq = V(Vinds{q},:); 
      U(q,:) = (Vq'*Vq + lambI) \ (Vq' * M(q,Vinds{q})');
    end
    parfor q=1:n
      Uq = U(Uinds{q},:); 
      V(q,:) = (Uq'*Uq + lambI) \ (Uq' * M(Uinds{q},q));
    end
  end
end
\end{lstlisting}
\begin{lstlisting}[language=Scala, firstline=9]
object BroadcastALS {
  def train(trainData: MLTable, k: Int, lambda: Double, 
	maxIter: Int): (LocalMatrix, LocalMatrix) = {
    val ctx = trainData.context
    val m = trainData.numRows
    val n = trainData.numCols
    val trainDataTrans = trainData.transpose
    val lambI = LocalMatrix.eye(k) * lambda
    // Initialize U and V matrices randomly
    val U0 = LocalMatrix.rand(m, k)
    val V0 = LocalMatrix.rand(n, k)
    (0 until maxIter).foldLeft((U0, V0))((UV, iterNum) => {
      val U = UV._1
      val V = UV._2
      // Broadcast V
      val V_b = ctx.broadcast(V)
      // Update U matrix
      val newU = computeFactor(trainData, V_b, lambI)
      // Broadcast U
      val U_b = ctx.broadcast(newU)
      // Update V matrix
      val newV = computeFactor(trainDataTrans, U_b, lambI)
      (newU, newV)
    })
  }

  def computeFactor(trainData: MLTable, fixedFactor: Broadcast[LocalMatrix], 
	lambI: LocalMatrix): LocalMatrix = {
    trainData.map(localALS(_, fixedFactor.value, lambI)).toLocalMatrix
  }

  def localALS(trainDataPart: MLRow, Y: LocalMatrix, lambI: LocalMatrix) = {
    val tuple = trainDataPart.tuple
    val Yq = Y.getRows(tuple.nonZeroIndices)
    val resultMat = ((Yq.transpose times Yq) + lambI).solve(Yq.transpose times tuple.nonZeroProjection)
    resultMat.toVector
  }
}
\end{lstlisting}
\caption{Matrix Factorization via ALS code in MATLAB (top) and \mlint (bottom).}
\label{code:bals}
\end{figure*}

\end{document}